\documentclass[review, sort&compress]{arxiv}

\usepackage[]{cite}
\usepackage{natbib,hyperref}
\usepackage{array}
\usepackage{amssymb,amsmath,subcaption}
\usepackage{graphicx}\graphicspath{{figs/}{eps/}}
\usepackage{psfrag}
\usepackage{ulem}
\usepackage{algorithmic}
\usepackage{algorithm}
\usepackage{varioref}
\usepackage{booktabs}
\usepackage{tabularx}
\journal{ArXiV}

\newcommand{\mat}[1]{\left[\begin{array}{#1}}

\usepackage{url}
\begin{document}

\begin{frontmatter}

\title{Learning a hyperplane regressor by minimizing an exact bound on the VC dimension\footnote{For commercial use or licensing of the MCM and its variants, contact the Foundation for Innovation and Technology Transfer, IIT Delhi.}}
\address[ee]{Department of Electrical Engineering}
\address[ma]{Department of Mathematics}
\address[cse]{Department of Computer Science \& Engineering}


\author[ee]{Jayadeva,~\textit{Senior Member,~IEEE}}
\ead{jayadeva@ee.iitd.ac.in}
\author[ma]{Suresh Chandra}
\author[ee]{Siddarth Sabharwal}
\author[cse]{Sanjit S. Batra}
\address{Indian Institute of Technology, Delhi}
\begin{abstract}
The capacity of a learning machine is measured by its Vapnik-Chervonenkis dimension, and learning machines with a low VC dimension generalize better. It is well known that the VC dimension of SVMs can be very large or unbounded, even though they generally yield state-of-the-art learning performance. In this paper, we show how to learn a hyperplane regressor by minimizing an exact, or \boldmath{$\Theta$} bound on its VC dimension. The proposed approach, termed as the Minimal Complexity Machine (MCM) Regressor, involves solving a simple linear programming problem. Experimental results show, that on a number of benchmark datasets, the proposed approach yields regressors with error rates much less than those obtained with conventional SVM regresssors, while often using fewer support vectors. On some benchmark datasets, the number of support vectors is less than one tenth the number used by SVMs, indicating that the MCM does indeed learn simpler representations.

\end{abstract}

\end{frontmatter}

\section{Introduction}
%
%

%
%
%
%
Support vector machines are amongst the most widely used machine learning techniques today. However, in his widely read tutorial, Burges \citep{burges1998} states that SVMs can have a very large Vapnik-Chervonenkis dimension, and that \textit{``at present there exists no theory which shows that good generalization performance is guaranteed for SVMs''}. The Vapnik-Chervonenkis (VC) dimension measures the capacity of a learning machine, and computational learning theory \citep{shawe1996framework, shawetaylor98, vapnik98, scholkopf2002learning} shows that a small VC dimension leads to good generalization. In recent work \citep{mcmneucom, mcmarxiv}, we have shown how to learn a classifier, termed as the Minimal Complexity Machine (MCM) classifier, that has a small VC dimension. The MCM classifier learns extremely sparse representations, often using less than one-tenth the number of support vectors used by SVMs, while yielding better test set accuracies.\\

This paper shows how to learn a hyperplane regressor, termed as the Minimal Complexity Machine (MCM) Regressor, by minimizing an exact (\boldmath{$\Theta$}) bound on the VC dimension. An exact bound implies that the proposed objective bounds the VC dimension from both above and below, and that the two are ``close''.  The MCM regressor requires the solution of a simple linear programming problem. Experimental results provided in the sequel show that the Minimal Complexity Machine outperforms conventional SVMs in terms of mean squared error on the test set, while often using far fewer support vectors. That the approach minimizes the machine capacity may be guaged from the fact that on many datasets, the MCM yields lower mean squared error (MSE) on test sets, while using less than $\frac{1}{10}$-th the number of support vectors obtained by SVMs.

The rest of the paper is organized as follows. Section \ref{linmcm} briefly describes the MCM classifier, for the sake of completeness. Section \ref{lmcrm} shows how to extend the approach to regression. Section \ref{kmcrm} discusses the extension of the Minimal Complexity Regression Machine to the kernel case. Section \ref{experimental} is devoted to a discussion of results obtained on selected benchmark datasets. Section \ref{conclusion} contains concluding remarks.

\section{The Linear Minimal Complexity Machine Classifier} \label{linmcm}
The motivation for the MCM originates from some sterling work on generalization \citep{shawe1996framework, shawetaylor98, vapnik98, scholkopf2002learning}. Consider a binary classification dataset with $n$-dimensional samples $x^i, i = 1, 2, ..., M$, where each sample is associated with a label $y_i \in \{+1, -1\}$. Vapnik \citep{vapnik98} showed that the VC dimension $\gamma$ for fat margin hyperplane classifiers with margin $d \geq d_{min}$ satisfies
\begin{equation}\label{eqnh}
\gamma \leq 1 + \operatorname{Min}(\frac{R^2}{d^2_{min}}, n)
\end{equation}
where $R$ denotes the radius of the smallest sphere enclosing all the training samples. Burges, in \citep{burges1998}, stated that \textit{``the above arguments strongly suggest that algorithms that minimize $\frac{R^2}{d^2}$ can be expected to give better generalization performance. Further evidence for this is found in the following theorem of (Vapnik, 1998), which we quote without proof''}.\\

Following this line of argument leads us to the formulations for a hyperplane classifier with minimum VC dimension; we term the same as the MCM classifier. We now summarize the MCM classifier formulation for the sake of completeness. Details may be found in \citep{mcmneucom, mcmarxiv}.

Consider the case of a linearly separable dataset. By definition, there exists a hyperplane that can classify these points with zero error. Let the separating hyperplane be given by
\begin{equation}
 u^Tx + v = 0.
\end{equation}

Let us denote
\begin{gather}
 h = \frac{\operatorname*{Max}_{i = 1, 2, ..., M} \;y_i(u^T x^i + v)}{\operatorname*{Min}_{i = 1, 2, ..., M} \;y_i(u^T x^i + v)}.
\end{gather}
In \citep{mcmneucom, mcmarxiv}, we show that there exist constants $\alpha, \beta > 0$, $\alpha, \beta \in \Re$ such that
\begin{equation}\label{exactbound}
 \alpha h^2 \leq \gamma \leq \beta h^2,
\end{equation}
or, in other words, $h^2$ constitutes a tight or exact ($\theta$) bound on the VC dimension $\gamma$. An exact bound implies that $h^2$ and $\gamma$ are close to each other.\\

\begin{figure}[hbtp]\label{fig1}
        \centering	
                \includegraphics[scale=0.5]{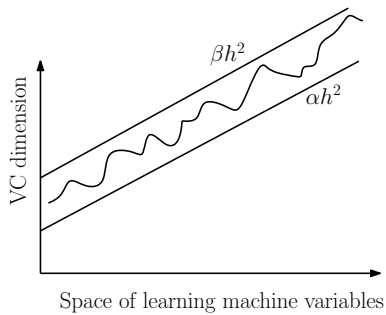}
                \caption{Illustration of the notion of an exact bound on the VC dimension. Even though the VC dimension $\gamma$ may have a complicated dependence on the variables defining the learning machine, multiples of $h^2$ bound $\gamma$ from both above and below. The exact bound $h^2$ is thus always ``close'' to the VC dimension, and minimizing $h^2$ with respect to the variables defining the learning machine allows us to find one that has a small VC dmension. }
\end{figure}

Figure \ref{fig1} illustrates this notion. The VC dimension is possibly related to the free parameters of a learning machine in a very complicated manner. It is known that the number of degrees of freedom in a learning machine is related to the VC dimension, but the connection is tenuous and usually abstruse. The use of a continuous and differentiable exact bound on the VC dimension allows us to find a learning machine with small VC dimension; this may be achieved by minimizing $h$ over the space of variables defining the separating hyperplane.\\

The MCM classifier solves an optimization problem, that tries to minimize the machine capacity, while classifying all training points of the linearly separable dataset correctly. This problem is given by
\begin{equation}\label{minh1}
\operatorname*{Minimize  }_{u, v} \; h ~=~ \frac{\operatorname*{Max}_{i = 1, 2, ..., M} \; y_i(u^T x^i + v)}{\operatorname*{Min}_{i = 1, 2, ..., M} \; y_i(u^T x^i + v)},
\end{equation}
that attempts to minimize $h$ instead of $h^2$, the square function $(\cdot)^2$ being a monotonically increasing one.

In \citep{mcmneucom, mcmarxiv}, we further show that the optimization problem (\ref{minh1}) may be reduced to the problem
\begin{gather}
\operatorname*{Min}_{w, b, h} ~~h \label{objm4}\\
h \geq y_i \cdot [{w^T x^i + b}], ~i = 1, 2, ..., M \label{consm41}\\
y_i \cdot [{w^T x^i + b}] \geq 1, ~i = 1, 2, ..., M, \label{consm42}
\end{gather}
where $w \in \Re^n$, and $b, h \in \Re$. We refer to the problem (\ref{objm4}) - (\ref{consm42}) as the hard margin Linear Minimum Complexity Machine (Linear MCM). \\

Once $w$ and $b$ have been determined by solving (\ref{objm4})-(\ref{consm42}), the class of a test sample $x$ may be determined from the sign of the discriminant function
\begin{equation}\label{testresult}
 f(x) = w^T x + b
\end{equation}

%

\section{The Linear Minimal Complexity Machine for Regression}\label{lmcrm}

In order to extend the MCM to regression, we refer to the work of Bi and Bennet \citep{bi2003geometric} that shows a link between the classification and regression tasks. Consider a regression problem with data points $x^i, i = 1, 2, ..., M$, and where the value of an unknown function at the point $x^i$ is denoted by $y_i \in \Re$. As before, we assume that the dimension of the input samples is $n$, i.e. $x^i = (x_1^i, x_2^i, ..., x_n^i)^T$. We address the task of $\epsilon$-regression, where $\epsilon$ is a tolerance parameter. It is required for the regressor to lie between the bounds ($y_i - \epsilon$) and ($y_i + \epsilon$) at the point $x^i$. It is easy to see that the task of building a regressor on this data has a one-to-one correspondence with a binary classification task in which class (-1) points lie at the $(n+1)$-dimensional co-ordinates ${(x^1; ~ y_1 - \epsilon), (x^2; ~ y_2 - \epsilon), ..., (x^M; ~ y_M - \epsilon)}$, and class (+1) points lie at the co-ordinates ${(x^1; ~ y_1 + \epsilon), (x^2; ~ y_2 + \epsilon), ..., (x^M; ~ y_M + \epsilon)}$. 

We first assume that these set of points are linearly separable, and we learn the classifier that separates the above training points. Let the separating hyperplane be given by $w^T x + \eta y + b = 0$. Then, the regressor is given by
\begin{equation}\label{testreg}
y = - \frac{1}{\eta} (w^T x + b)
\end{equation}

Following (\ref{objm4}) - (\ref{consm42}), the hard margin linear MCM classifier for the above $(n+1)$-dimensional samples is obtained by solving the optimization problem

\begin{gather}
\operatorname*{Min}_{w, b, h} ~~h \label{obj5}\\
h \geq 1 \cdot [(w^T x^i + b) + \eta (y_i + \epsilon)] , ~i = 1, 2, ..., M \label{cons51}\\
1 \cdot [(w^T x^i + b) + \eta (y_i + \epsilon)] \geq 1, ~i = 1, 2, ..., M \label{cons52}\\
h \geq -1 \cdot [(w^T x^i + b) + \eta (y_i - \epsilon)] , ~i = 1, 2, ..., M \label{cons53}\\
-1 \cdot [(w^T x^i + b) + \eta (y_i - \epsilon)] \geq 1, ~i = 1, 2, ..., M \label{cons54}
\end{gather}
Observe that the first two constraints, given by (\ref{cons51}) and (\ref{cons52}), correspond to class (+1) samples; the multiplier (+1) in (\ref{cons51}) and (\ref{cons52}) corresponds to samples with $y_i = 1$ in (\ref{consm41}) and (\ref{consm42}). Similarly, constraints (\ref{cons53}) and (\ref{cons54}) correspond to class (-1) samples; the multiplier (-1) in (\ref{cons53}) and (\ref{cons54}) corresponds to samples with $y_i = -1$ in (\ref{consm41}) and (\ref{consm42}). After solving (\ref{obj5})-(\ref{cons54}), we obtain $w$ and $b$. The regressor is given by (\ref{testreg}).\\

$\cdot$ Note that the notion of the VC dimension has not been explored in the case of regression. The work of Bi and Bennett allows us to do just that by using the notion of an exact bound on the VC dimension of a classifier.\\

In practice, the regressor will not lie strictly between the samples $\{x^i; y_i + \epsilon, \; i = 1, 2,..., M\}$ and $\{x^i; y_i - \epsilon, \; i = 1, 2,..., M\}$, and the optimization problem needs to be modified to allow for a trade-off between the empirical error and the VC dimension bound. We elaborate on the soft margin MCM regressor in section \ref{softmcrm}.

\subsection{The soft margin linear MCM regressor}\label{softmcrm}

The soft margin MCM regressor is obtained by solving the following optimization problem.


\begin{gather}
\operatorname*{Min}_{w, b, h, q} ~~h ~+~ C \cdot \sum_{i = 1}^M (q_i^{+} + q_i^{-}) \label{obj9}\\
h \geq 1 \cdot [(w^T x^i + b) + \eta (y_i + \epsilon)] , ~i = 1, 2, ..., M \label{cons91}\\
[(w^T x^i + b) + \eta (y_i + \epsilon)] + q_i^{+} \geq 1, ~i = 1, 2, ..., M \label{cons92}\\
h \geq -1 \cdot [(w^T x^i + b) + \eta (y_i - \epsilon)] , ~i = 1, 2, ..., M \label{cons93}\\
-1 \cdot [(w^T x^i + b) + \eta (y_i - \epsilon)] + q_i^{-} \geq 1, ~i = 1, 2, ..., M \label{cons94}\\
q_i^+, q_i^- \geq 0, ~i = 1, 2, ..., M \label{cons95}
\end{gather}

Here, $C$ determines the trade-off between the VC dimension bound and the empirical error.

In order to regress more complex functions, we extend the MCM regressor in the next section to the kernel case.

\section{The Kernel MCM Regressor}\label{kmcrm}
We consider a map $\phi(x)$ that maps the input samples from $\Re^n$ to $\Re^d$, where $d > n$. Corresponding to (\ref{testreg}), the hyperplane in the image space is  given by

\begin{equation}
 y = - \frac{1}{\eta} (w^T \phi(x) + b)
\end{equation}

Following (\ref{obj9}) - (\ref{cons94}), the corresponding optimization problem for the kernel MCM may be shown to be
\begin{gather}
\operatorname*{Min}_{w, b, h, q} ~~h ~+~ C \cdot \sum_{i = 1}^M (q_i^{+} + q_i^{-}) \label{kobj9}\\
h \geq 1 \cdot [(w^T \phi(x^i) + b) + \eta (y_i + \epsilon)] , ~i = 1, 2, ..., M \label{consk91}\\
[(w^T \phi(x^i) + b) + \eta (y_i + \epsilon)] + q_i^{+} \geq 1, ~i = 1, 2, ..., M \label{consk92}\\
h \geq -1 \cdot [(w^T \phi(x^i) + b) + \eta (y_i - \epsilon)] , ~i = 1, 2, ..., M \label{consk93}\\
-1 \cdot [(w^T \phi(x^i) + b) + \eta (y_i - \epsilon)] + q_i^{-} \geq 1, ~i = 1, 2, ..., M \label{consk94}\\
q_i^+, q_i^- \geq 0, ~i = 1, 2, ..., M \label{consk95}
\end{gather}


The image vectors $\phi(x^i), i = 1, 2, ..., M$ form an overcomplete basis in the empirical feature space, in which $w$ also lies. Hence, we can write
\begin{equation}
 w = \sum_{j = 1}^M \lambda_j \phi(x^j).
\end{equation}\label{weqsumlambda}

Therefore,
\begin{gather}\label{simpliphi}
 w^T \phi(x^i) + b = \sum_{j = 1}^M \lambda_j \phi(x^j)^T\phi(x^i) + b = \sum_{j = 1}^M \lambda_j K(x^i, x^j) + b,
\end{gather}

where $K(p, q)$ denotes the Kernel function with input vectors $p$ and $q$, and is defined as
\begin{equation}
 K(p, q) = \phi(p)^T \phi(q).
\end{equation}\label{kernel}

Substituting from (\ref{simpliphi}) into (\ref{kobj9}) - (\ref{consk94}), we obtain the following optimization problem.

\begin{gather}
\operatorname*{Min}_{w, b, h, q} ~~h ~+~ C \cdot \sum_{i = 1}^M (q_i^{+} + q_i^{-}) \label{kobj10}\\
h \geq 1 \cdot [(\sum_{j = 1}^M \lambda_j K(x^i, x^j) + b) + \eta (y_i + \epsilon)] , ~i = 1, 2, ..., M \label{consk101}\\
[(\sum_{j = 1}^M \lambda_j K(x^i, x^j) + b) + \eta (y_i + \epsilon)] + q_i^{+} \geq 1, ~i = 1, 2, ..., M \label{consk102}\\
h \geq -1 \cdot [(\sum_{j = 1}^M \lambda_j K(x^i, x^j) + b) + \eta (y_i - \epsilon)] , ~i = 1, 2, ..., M \label{consk103}\\
-1 \cdot [(\sum_{j = 1}^M \lambda_j K(x^i, x^j) + b) + \eta (y_i - \epsilon)] + q_i^{-} \geq 1, ~i = 1, 2, ..., M \label{consk104}\\
q_i^+, q_i^- \geq 0, ~i = 1, 2, ..., M \label{consk105}
\end{gather}


Once the variables $\lambda_j, j = 1, 2, ..., M$ and $\hat{b}$ have been determined by solving (\ref{kobj10})-(\ref{consk105}), the value of the regressor at $x$ is given by
\begin{equation}
y = -\frac{1}{\eta} (\sum_{j = 1}^M \lambda_j K(x, x^j) + b)
\end{equation}

\section{Experimental results}\label{experimental}
The MCM regression formulations were coded in MATLAB. A MCM tool is under development and would shortly be available from the homepage of the first author. Table \ref{table1} summarizes five fold cross validation results of the linear MCM regressor on a number of datasets taken from the UCI machine learning repository. The table indicates the mean squared errors for the linear MCM and classical SVM regressor.


\begin{table}[htbp]
\caption{Linear MCM regression results}
\begin{tabular}{|c|c|c|}
\hline
 & \multicolumn{ 2}{c|}{Mean Squared Error} \\ \hline
Dataset (Dimensions) & MCM & SVM \\ \hline
 &  &  \\ \hline
Autompg (398$\times$8) & 0.35 $\pm$ 0.02 & 0.36 $\pm$ 0.03 \\ \hline
 &  &  \\ \hline
Yacht (308$\times$7) & 104.8 $\pm$ 7.5 & 161.8 $\pm$ 68.4 \\ \hline
 &  &  \\ \hline
Price (159$\times$16) & 33.6 $\pm$ 12.5 (in million dollars) & 32.8 $\pm$ 23.2 (in million dollars) \\ \hline
 &  &  \\ \hline
Machine (209$\times$7) & 6.5368 $\pm$ 3.6512 (thousand units) & 19.948 $\pm$ 15.521 (thousand units) \\ \hline
 &  &  \\ \hline
Baseball (337$\times$17) & 0.80 $\pm$ 0.12 (in million dollars) & 1.62 $\pm$ 0.61 (in million dollars) \\ \hline
 &  &  \\ \hline
Housing (506$\times$13) & 23.09 $\pm$ 4.26 & 25.92 $\pm$ 9.61 \\ \hline
 &  &  \\ \hline
Energy Efficiency (768$\times$8) & 8.74 $\pm$ 1.35 & 9.08 $\pm$ 1.45 \\ \hline
\end{tabular}
\label{table1}
\end{table}

  Accuracies are indicated as mean $\pm$ standard deviation, computed over the five folds in a cross validation setting. The table compares the linear MCM with LIBSVM using a linear kernel. The values of $C$ were determined for the MCM by performing a grid search.\\


\begin{table}[htbp]
\centering
\footnotesize\setlength{\tabcolsep}{2.5pt}
\caption{Kernel MCM regression results}
\begin{tabular}{|c|c|c|c|c|}
\hline
 & \multicolumn{ 2}{c|}{Mean Squared Error} & \multicolumn{ 2}{c|}{Number of Support Vectors} \\ \hline
Dataset & MCM & SVM & MCM & SVM \\ \hline
 &  &  &  &  \\ \hline
Autompg & 0.31 $\pm$ 0.02 & 0.32 $\pm$ 0.04 & 26.8 $\pm$ 7.9 & 184.2 $\pm$ 4.3 \\ \hline
 &  &  &  &  \\ \hline
Yacht & 0.97 $\pm$ 0.42 & 158.86 $\pm$ 62.9 & 129.8 $\pm$ 24.3 & 224.8 $\pm$ 0.8 \\ \hline
 &  &  &  &  \\ \hline
Price & 12.77 $\pm$ 9.0 (mill. \$) & 39.48 $\pm$ 26.9 (mill. \$) & 68.6 $\pm$ 15.4 & 126.4 $\pm$ 0.9 \\ \hline
 &  &  &  &  \\ \hline
Machine & 7.588 $\pm$ 3.909 (th. units) & 26.351 $\pm$ 21.330 (th. units) & 52.4 $\pm$ 27.3 & 166.4 $\pm$ 1.5 \\ \hline
 &  &  &  &  \\ \hline
Baseball & 0.78 $\pm$ 0.14 (mill. \$) & 1.78 $\pm$ 0.67 (mill. \$) & 24.4 $\pm$ 6.8 & 269.2 $\pm$ 1.1 \\ \hline
 &  &  &  &  \\ \hline
Housing & 25.8 $\pm$ 4.64 & 29.72 $\pm$ 5.96 & 76.4 $\pm$ 14.45 & 386.8 $\pm$ 4.82 \\ \hline
 &  &  &  &  \\ \hline
Energy Efficiency & 4.1 $\pm$ 0.2 & 7.64 $\pm$ 1.31 & 44 $\pm$ 3.39 & 557 $\pm$ 5.05 \\ \hline
\end{tabular}
\label{table2}
\end{table}

Table \ref{table2} summarizes five fold cross validation results of the kernel MCM regressor on a number of datasets. The width of the Gaussian kernel was chosen by using a grid search. The table shows the mean squared error and the number of support vectors for both the kernel MCM and the classical SVM with a Gaussian kernel. The results indicate that the kernel MCM yields better generalization than the SVM. In the case of kernel regression, the MCM uses fewer support vectors - note that in the case of some of the datasets, the MCM uses \textit{less than one-tenth} the number of support vectors required by a SVM. The large difference with the SVM results indicates that despite good performance, SVMs may still be far from the optimal solution.
\\

Vapnik \citep{vapnik95} showed that
\begin{equation}
E(P_{error}) \leq \frac{E(\# \mbox{support vectors})}{\# \mbox{training samples}},
\end{equation}\label{vcbound2}
where $E(P_{error})$ denotes the expected test set error, $\# \mbox{training samples}$  denotes the number of training samples, and the expected number of support vectors obtained on training sets of the same size is denoted by $E(\#$ support vectors $)$. Thus, the results support the claim of minimizing the VC dimension of the learning machine.\\
We also observe that the variance of the number of support vectors is large in Table \ref{table2}. This is in keeping with the recent work of \citep{xu2012sparse} which shows that sparse algorithms tend to vary a lot with changes in the training data.

\section{Conclusion} \label{conclusion}
In this paper, we propose a way to build a hyperplane regressor, termed as the Minimal Complexity Machine (MCM) regressor, that attempts to minimize an exact bound on the VC dimension. The regressor can be found by solving a linear programming problem. Experimental results show that the regressor outperforms the classical SVM regressor in terms of test set error on many selected benchmark datasets. The number of support vectors is less in the case of the MCM, often by a substantial factor, in comparison to the classical SVM. It has not escaped our attention that the proposed approach can be extended to least squares regression, as well as to other tasks; in fact, a large number of variants of SVMs can be re-examined with the objective of minimizing the VC dimension. The MCM can also be incorporated into frameworks such as DTMKL \citep{duan2012domain} to tackle cross-domain learning problems.

\bibliographystyle{plainnat}
\bibliography{linear-struct-min3-reg-arxiv}

\begin{thebibliography}{11}
\providecommand{\natexlab}[1]{#1}
\providecommand{\url}[1]{\texttt{#1}}
\expandafter\ifx\csname urlstyle\endcsname\relax
  \providecommand{\doi}[1]{doi: #1}\else
  \providecommand{\doi}{doi: \begingroup \urlstyle{rm}\Url}\fi

\bibitem[Bi and Bennett(2003)]{bi2003geometric}
Jinbo Bi and Kristin~P Bennett.
\newblock A geometric approach to support vector regression.
\newblock \emph{Neurocomputing}, 55\penalty0 (1):\penalty0 79--108, 2003.

\bibitem[Burges(1998)]{burges1998}
Christopher~JC Burges.
\newblock A tutorial on support vector machines for pattern recognition.
\newblock \emph{Data mining and knowledge discovery}, 2\penalty0 (2):\penalty0
  121--167, 1998.

\bibitem[Cortes and Vapnik(1995)]{vapnik95}
Corinna Cortes and Vladimir Vapnik.
\newblock Support-vector networks.
\newblock \emph{Machine learning}, 20\penalty0 (3):\penalty0 273--297, 1995.

\bibitem[Duan et~al.(2012)Duan, Tsang, and Xu]{duan2012domain}
Lixin Duan, Ivor~W Tsang, and Dong Xu.
\newblock Domain transfer multiple kernel learning.
\newblock \emph{Pattern Analysis and Machine Intelligence, IEEE Transactions
  on}, 34\penalty0 (3):\penalty0 465--479, 2012.

\bibitem[{Jayadeva}(2014)]{mcmarxiv}
{Jayadeva}.
\newblock {Learning a hyperplane classifier by minimizing an exact bound on the
  VC dimension}.
\newblock \emph{ArXiv e-prints}, August 2014.

\bibitem[Jayadeva(2014)]{mcmneucom}
Jayadeva.
\newblock Learning a hyperplane classifier by minimizing an exact bound on the
  \{VC\} dimension.
\newblock \emph{Neurocomputing}, \penalty0 (0):\penalty0 --, 2014.
\newblock ISSN 0925-2312.
\newblock \doi{http://dx.doi.org/10.1016/j.neucom.2014.07.062}.
\newblock URL
  \url{http://www.sciencedirect.com/science/article/pii/S0925231214010194}.

\bibitem[Sch{\"o}lkopf and Smola(2002)]{scholkopf2002learning}
Bernhard Sch{\"o}lkopf and Alexander~J Smola.
\newblock \emph{Learning with kernels}.
\newblock “The” MIT Press, 2002.

\bibitem[Shawe-Taylor et~al.(1996)Shawe-Taylor, Bartlett, Williamson, and
  Anthony]{shawe1996framework}
John Shawe-Taylor, Peter~L Bartlett, Robert~C Williamson, and Martin Anthony.
\newblock A framework for structural risk minimisation.
\newblock In \emph{Proceedings of the ninth annual conference on Computational
  learning theory}, pages 68--76. ACM, 1996.

\bibitem[Shawe-Taylor et~al.(1998)Shawe-Taylor, Bartlett, Williamson, and
  Anthony]{shawetaylor98}
John Shawe-Taylor, Peter~L Bartlett, Robert~C Williamson, and Martin Anthony.
\newblock Structural risk minimization over data-dependent hierarchies.
\newblock \emph{Information Theory, IEEE Transactions on}, 44\penalty0
  (5):\penalty0 1926--1940, 1998.

\bibitem[Vapnik(1998)]{vapnik98}
Vladimir~N Vapnik.
\newblock Statistical learning theory.
\newblock 1998.

\bibitem[Xu et~al.(2012)Xu, Caramanis, and Mannor]{xu2012sparse}
Huan Xu, Constantine Caramanis, and Shie Mannor.
\newblock Sparse algorithms are not stable: A no-free-lunch theorem.
\newblock \emph{Pattern Analysis and Machine Intelligence, IEEE Transactions
  on}, 34\penalty0 (1):\penalty0 187--193, 2012.

\end{thebibliography}

\end{document}